\pdfoutput=1
\documentclass[letterpaper]{article} 
\usepackage{aaai24}  
\usepackage{times}  
\usepackage{helvet}  
\usepackage{courier}  
\usepackage[hyphens]{url}  
\usepackage{graphicx} 
\usepackage{natbib}  
\usepackage{caption} 
\usepackage{algorithm}
\usepackage{algorithmic}

\usepackage{hyperref}
\usepackage{times}
\usepackage{latexsym}
\usepackage{caption}
\usepackage{subcaption}
\usepackage{soul}
\usepackage{xcolor}
\usepackage{tabularx}
\usepackage{multirow}
\usepackage{booktabs}
\usepackage{microtype}
\usepackage{amsmath,amssymb,amsthm}
\usepackage{mathrsfs}
\usepackage{inconsolata}
\usepackage{microtype}
\usepackage{graphics}
\usepackage{graphicx}
\usepackage{subcaption}
\usepackage[flushleft]{threeparttable}
\usepackage{fdsymbol}

\usepackage{newfloat}
\usepackage{listings}
\DeclareCaptionStyle{ruled}{labelfont=normalfont,labelsep=colon,strut=off} 
\floatstyle{ruled}
\newfloat{listing}{tb}{lst}{}
\floatname{listing}{Listing}
\usepackage[utf8]{inputenc}
\def \bb{\textit{B}}
\def \ss{\mathbf{s}}
\title{BBScore: A Brownian Bridge Based Metric for Assessing Text Coherence}
\author{
    Zhecheng Sheng\equalcontrib, Tianhao Zhang\equalcontrib, Chen Jiang\equalcontrib, Dongyeop Kang \\
}
\affiliations{
    University of Minnesota, Twin Cities\\

    Minneapolis, Minnesota 55455\\
    \{sheng136, zhan7594, jian0649, dongyeop\}@umn.edu

}

\usepackage{bibentry}

\begin{document}

\maketitle

\begin{abstract}
Measuring the coherence of text is a vital aspect of evaluating the quality of written content. Recent advancements in neural coherence modeling have demonstrated their efficacy in capturing entity coreference and discourse relations, thereby enhancing coherence evaluation. However, many existing methods heavily depend on static embeddings or focus narrowly on nearby context, constraining their capacity to measure the overarching coherence of long texts.
In this paper, we posit that coherent texts inherently manifest a sequential and cohesive interplay among sentences, effectively conveying the central theme, purpose, or standpoint. To explore this abstract relationship, we introduce the "BBScore," a novel reference-free metric grounded in Brownian bridge theory for assessing text coherence. Our findings showcase that when synergized with a simple additional classification component, this metric attains a performance level comparable to state-of-the-art techniques on standard artificial discrimination tasks.
We also establish in downstream tasks that this metric effectively differentiates between human-written documents and text generated by large language models under a specific domain. Furthermore, we illustrate the efficacy of this approach in detecting written styles attributed to diverse large language models, underscoring its potential for generalizability. In summary, we present a novel Brownian bridge coherence metric capable of measuring both local and global text coherence, while circumventing the need for end-to-end model training. This flexibility allows for its application in various downstream tasks. The code for calculating BBScore is available at \url{https://github.com/zcsheng95/BBScore}. 
\end{abstract}

\section{Introduction}

Text coherence is the quality of a written or spoken text that makes it internally consistent, logical, and easily understandable to the reader or listener. 
It encompasses the flow and connectivity of ideas, the logical progression of information, and the smooth transition between sentences and paragraphs. 
Evaluating the coherence of text requires the evaluator to differentiate between a set of individual utterances and naturally flowing discourses such as co-references or grammatical relationships \citep{wang2014short,kehler2022coherence}.

In linguistic theory, the evaluation of text coherence can be performed at both the local and global levels \citep{agar1982interpreting, glosser1992comparison, kintsch1978toward}. Local coherence focuses on the interrelation of information within a specific unit of discourse and its connection to the preceding unit. Conversely, global coherence refers to the degree to which the various components of discourse maintain a cohesive representation of the main topic. 
Despite the emergence of neural coherence models, most prior works focused on static representations of their local neighboring texts. \citep{tien-nguyen-joty-2017-neural, mesgar-strube-2018-neural, mesgar-etal-2021-neural-graph, jeon-strube-2022-entity}
\begin{figure}
    \centering
    \includegraphics[width=0.85\linewidth]{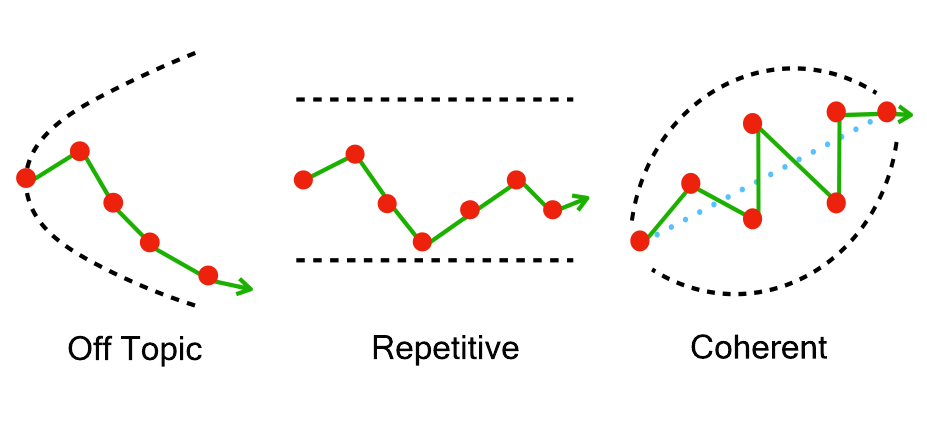}
    \caption{Utterance flow in latent space}
    \vspace{-4mm}
    \label{Figure:latents_flow}
\end{figure}

This paper aims to capture \textit{global coherence} from text, a sequential and cohesive relationship among the sentences or words, representing the main idea, goal, opinion, and other pertinent aspects regarding a coherent text. 
This relationship may vary across different situations, individuals, and even over time for the same individual. To simplify this concept and enhance its accessibility, we assume that this relationship remains consistent within a fixed domain or dataset. Also, we consider a coherent text should encompass the following properties: (1) A main idea should be kept throughout the text (2) Sentences near the beginning and ending part of the article should emphasize the main idea and (3) Sentences near the middle part can go a little bit far from the main idea, but still be controlled.

We present a novel metric that can capture global text coherence using the Brownian bridge framework, which has been used to analyze animal movements \citep{Horne:2007aa} or long-text generation \citep{wang2023language}.
Brownian bridges serve as continuous-time stochastic models for movement, where the probability of being in a particular area is conditioned on the starting and ending locations, the time elapsed between those points, and the mobility or speed of movement  \citep{Horne:2007aa,chow2009brownian}. 
More specifically, we assume each step of this movement can be viewed as a sentence or word within the text, and the trajectory of the movement can represent different levels of coherence. Three such examples are given in Figure \ref{Figure:latents_flow}.
In this paper, our contributions and findings are summarized as follows:
\begin{itemize}
    \item We introduce $\sigma_m$ in the Brownian bridge as a parameter that characterizes the text style.
    \item We design a flexible score measurement based on the Brownian bridge that can be used to evaluate the overall coherence of a text and naturally fit into any downstream tasks.
    \item We show our score can capture both global and local coherence and can effectively differentiate large language models generated and human-written text.
\end{itemize}

\section{Related Work}

Pre-neural models have predominantly relied on entity-based methods \citep{barzilay-lapata-2008-modeling, elsner-charniak-2011-extending} to model coherence in text, as they closely tie to linguistic theory. This class of methods builds upon the Centering theory \citep{grosz-etal-1995-centering}, which assumes that each sentence contains a centered entity that focuses the attention of the entire discourse. These methods provide insights into how information is connected at the local level but often fall short of capturing global coherence. With recent advancements in neural models such as the Transformer \citep{attention-2017}, efforts have been made to improve these methods using deep network architectures \citep{guinaudeau-strube-2013-graph, tien-nguyen-joty-2017-neural, adewoyin-etal-2022-rstgen}, integrating various lexical information to enhance performance. 

Previous studies have revealed that Transformer language models struggle to effectively capture coherence structures \citep{deng2022model}. To address this limitation and guide language models in learning text dynamics, representation learning methods have incorporated auto variational encoders \citep{bowman2016generating} or contrastive learning techniques \citep{gao2022simcse}, which aim to model sentence embeddings by leveraging information from neighboring utterances. However, these approaches yield static representations that overlook the dynamics of the entire text. In a recent study, \citet{wang2023language} proposed an innovative method utilizing the Brownian bridge to model long-range text dependencies. The proposed method exhibited superior performance in generating long and coherent discourses compared to the baseline GPT2 model. In this paper, we extend their findings by introducing a likelihood measurement and conducting experiments to assess its efficacy in capturing both global text coherence and local coherence.

The Shuffle test \citep{barzilay-lapata-2008-modeling} has been widely utilized as a common practice for assessing coherence models. This artificial task requires the model to distinguish between an original document and its randomly shuffled counterpart. \citet{laban-etal-2021-transformer} proposes that it should be treated as a probe—an evaluation task that allows models to be assessed without explicit supervision. Notably, they argue that the Shuffle test is not limited to local coherence analysis and can be extended to longer sequences, such as in the k-Blocked Shuffle test \citep{van1985semantic,hearst-1997-text}, where the block size can provide insights into a model's ability to discern global coherence or identify main topics and subtopics.

\section{Preliminary} \label{methods}
\paragraph{Brownian Bridge}
A Brownian bridge is a stochastic process that modifies the standard Brownian motion by requiring an extra constraint \citep{chow2009brownian,revuz2013continuous}. A standard Brownian bridge $\{B(t)|t\in [0,1]\}$ is a Gaussian stochastic process such that $E[B(t)] = 0$ for all $t\in[0,1]$ and $E[B(t_1)B(t_2)]=\min(t_1,t_2)-t_1t_2$ for all $t_1,t_2\in [0,1]$. 

However, in our paper, we modified the standard Brownian bridge with a constant called diffusion coefficient and denoted it as $\sigma_m^2$ to encode the information of the domain-specific property. Similar methods are also used to model animal mobility \citep{Horne:2007aa} and human mobility \citep{John:2021}, but to the best of our knowledge, it is the first to use this assumption for evaluating text coherence. Formally, this modified Brownian bridge $\{S(t)|t\in [0,T]\}$ follows a normal distribution $S(t)\sim \mathcal{N}(\mu(t),\sigma^2(t) \mathbf{I})$, for $t\in[0,T]$, where
\[
\footnotesize
\begin{aligned}
\mu(t) = a + \frac{t}{T} (b-a),\quad \sigma^2(t)= \frac{t(T-t)}{T} \sigma^2_m
\end{aligned}
.\]
$S(0)=a \in \mathbb{R}^n$ and $S(T)=b\in \mathbb{R}^n$ are the starting and ending location respectively, and $n\in \mathbb{N}$ is a fixed constant.
\paragraph{Brownian Encoder via Contrastive Learning}
To map utterances from the observable space into the designed latent space, we followed the practice from the previous work \citep{wang2023language}. The weights from pretrained GPT2 \citep{radford2019language} are frozen while at the top of its hidden states, a trainable multi-layer perceptron is appended.
The training details of the encoder are as follows. Given a sequence of data points $X=\{x_1,\dots,x_N\}$, we randomly draw batches $\mathcal{B}$ consisting of randomly sampled triplets $(x_{i_1}, x_{i_2}, x_{i_3})$ with $1\le i_1 < i_2 < i_3 \le T$ and optimize the encoder by minimize the corresponding contrastive objective function:
\begin{equation*}
\footnotesize
\begin{aligned}
\mathcal{L} =\mathbb{E}_X\left[-\log \frac{\exp(d(x_{i_1}, x_{i_2}, x_{i_3}; f_\theta))}{\sum\limits_{(x_{i_1}, x_{i_2'},x_{i_3})\in \mathcal{B}}\exp(d(x_{i_1}, x_{i_2'},x_{i_3}; f_\theta))}\right]
\end{aligned}
\end{equation*}
with $d(x_{i_1}, x_{i_2},x_{i_3}; f_\theta)$ defined by:
\begin{equation*}
\footnotesize
\begin{aligned}
-\frac{\|f_\theta(x_{i_2}) - (1-\frac{i_2-i_1}{i_3-i_1})f_\theta(x_{i_1}) -\frac{i_2-i_1}{i_3-i_1}f_\theta (x_{i_3})\|_2^2}{2\sigma^2},
\end{aligned}
\end{equation*}
where $f_\theta$ denotes the encoder and $\sigma^2=\frac{(i_2-i_1)(i_3-i_2)}{i_3-i_1}$. The loss function can be viewed as maximizing towards the triplets sampled from a Brownian bridge process while minimizing the function for triplets sampled from a different sequence. We implemented the encoder training step based on code from \cite{wang2023language}. As one can find that during the training, the diffusion coefficient $\sigma^2_m$ does not show up in the  $\sigma^2$, and instead, this information is encoded into the function, $f_\theta$. After the training is finished, we can approximate the diffusion coefficient $\sigma^2_m$ based on the training set and the encoder $f_\theta$ which will be discussed in the following section. 

\section{Proposed Methodology: Brownian Bridge Score for Text Coherence}

\begin{figure}[t!]
\centering
\includegraphics[width=0.9\linewidth]{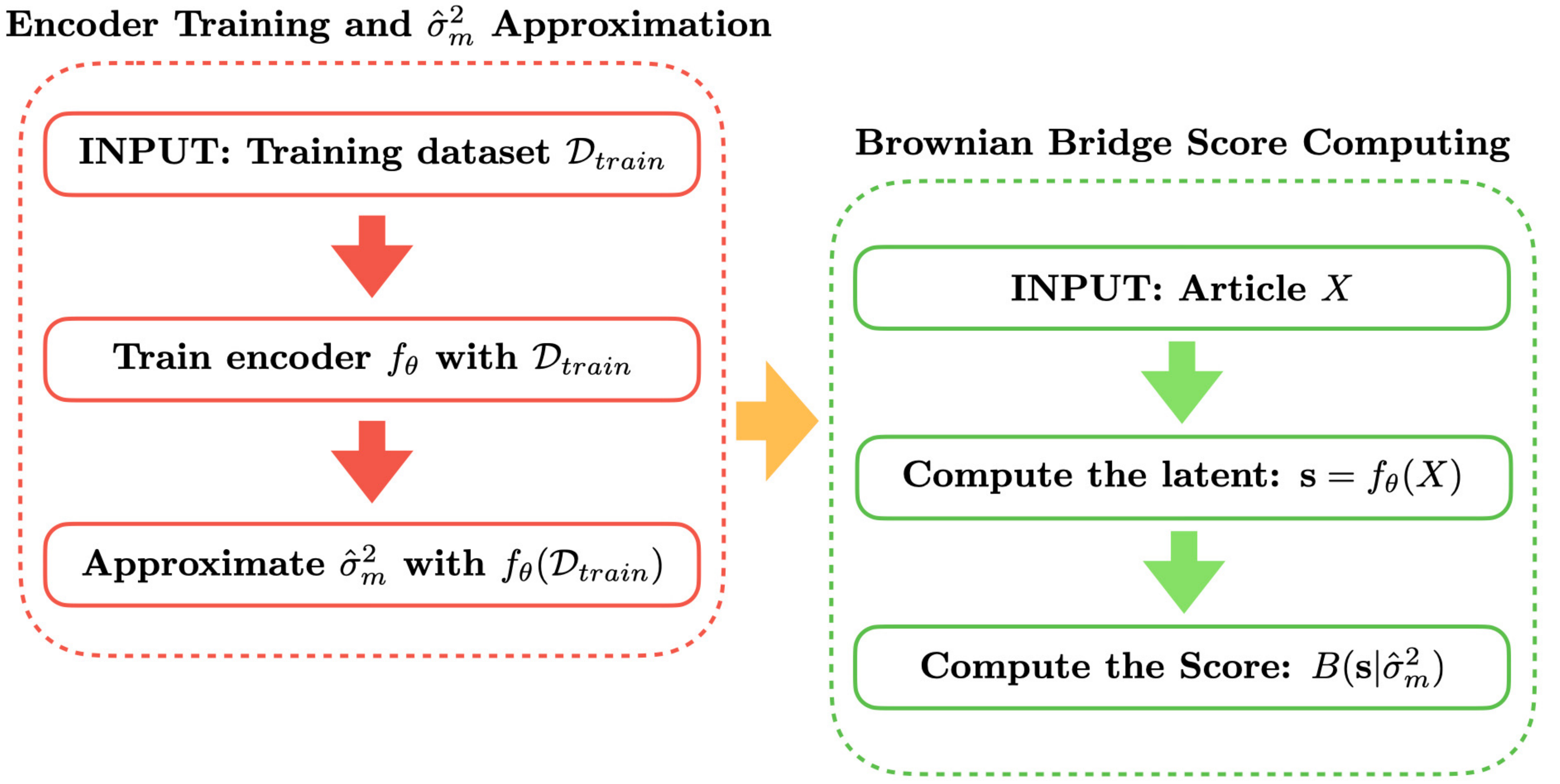}
\caption{Procedure to generate the Brownian bridge score, $\bb (\ss| \hat{\sigma}_m^2)$.}
\label{Figure:score_computation}
\end{figure}
In this section, we introduce how to approximate the diffusion coefficient and how to design our Brownian bridge score (BBScore). First of all, as shown in Figure \ref{Figure:score_computation}, we display the general procedure to use our BBScore: Given a domain-specific dataset $\mathcal{D}_{train}$ and train the encoder $f_\theta$ with this dataset. Then, approximate the diffusion coefficient $\sigma_m^2$ with all the latents $f_\theta(\mathcal{D}_{train})$. After this, we can compute the BBScore of any given article which is assumed to be in the same domain as the training dataset $\mathcal{D}_{train}$. The overall procedure to compute the BBScore is shown in the green dash box. Next, we will introduce how to approximate $\sigma_m^2$ and construct BBScore $\bb(\cdot | \sigma_m^2)$.

The approximation and construction are both inspired by the likelihood of Brownian bridge. Given a Brownian bridge sequence $\ss = (s_1,s_2,\cdots, s_{T(\ss)})$, where $s_i\in\mathbb{R}^n$, for $i=2,3,\cdots (T(\ss)-1)$ and $T(\ss)$ is the length of $\ss$, it satisfies: $s_i \sim \mathit{N}(\mu_i, \sigma_i^2\mathbf{I})$, where $\mathbf{I}$ is the $n\times n$ identity matrix and 
\begin{equation*}
\resizebox{0.9\linewidth}{!}{$
\mu_i = s_1 + \frac{i-1}{T(\ss)} (s_{T(\ss)} - s_1),\ \sigma^2_i = \frac{(i-1)(T(\ss)-i)}{T(\ss)} \sigma^2_m
$.}
\end{equation*}

To estimate $\sigma^2_m$ with the training dataset $\mathcal{D}_{train}$, we consider the likelihood: for $1<i<T(\ss)$, the likelihood of location $s_i$ is: 
\[
\footnotesize
\begin{aligned}
\mathcal{L}_i = \frac{1}{2\pi\sigma^2_i} \textrm{Exp}(-\frac{1}{2\sigma^2_i}||s_i - \mu_i||^2).
\end{aligned}
\]

Then the product of the log-likelihood of $\mathcal{L}_i$ is
\begin{equation}
\footnotesize 
\begin{aligned}
\label{eq:log_likelihood}
\ln (\prod_{i=2}^{T(\ss)-1} \mathcal{L}_i) = -\sum_{i=2}^{T(\ss)-1} (\ln(\alpha_i(\ss) \sigma_m^2) + \frac{\beta_i(\ss)}{\sigma_m^2}),
\end{aligned}
\end{equation}
where for $i=2,3,\cdots T(\ss)-1$
\begin{equation*}
\footnotesize
\begin{aligned}
\alpha_i(\ss) = \frac{2\pi (i-1)(T(\ss)-i)}{T(\ss)-1}, \beta_i(\ss) = \frac{(T(\ss)-1)||s_i -\mu_i||^2}{ 2(i-1) (T(\ss)-i)}.
\end{aligned}
\end{equation*}

Next, by setting the derivative of Eq (\ref{eq:log_likelihood}) to be zero, we can find the best approximation of $\sigma_m^2$ with maximum likelihood: $\hat{\sigma}_m^2(\ss) = \frac{1}{T(\ss)-2} \sum_{i=2}^{T(\ss)-1} \beta_i(\ss)$. Therefore, we can approximate the $\sigma_m^2$ with the training set: \[\hat{\sigma}_m^2 = \sum_{\ss \in f_\theta(\mathcal{D}_{train})} \hat{\sigma}_m^2(\ss).\]
And to further simplify our method, we use the average of $n$ dimension of the approximated $\hat{\sigma}_m^2$ as the diffusion coefficient used in the BBScore. Besides, we show the necessity of such approximation by a simple sensitivity analysis, e.g. the standard Brownian bridge $\sigma_m^2=1$ cannot capture the coherence of the article (See Appendix \ref{sec:app1_DCA}). Next, we define the BBScore which is inspired by the log-likelihood:
\begin{equation}
\label{eq:evaluation_measure}
\footnotesize
\begin{aligned}
\bb (\ss| \hat{\sigma}_m^2) = \frac{|\sum_{i=2}^{T(\ss)-1} \ln(\alpha_i(\ss) \hat{\sigma}_m^2)+ \frac{\beta_i(\ss)}{\hat{\sigma}_m^2}|}{T(\ss)-2}.
\end{aligned}
\end{equation}
Based on the likelihood (Eq \eqref{eq:log_likelihood}), this measure should show the following relation with text coherence: the more the text is coherent, the smaller the value of $\bb (\ss| \hat{\sigma}_m^2)$ will be. 

\section{Experiments}
\subsection{Research Questions}
Throughout our experiments, we address two research questions as follows:
\begin{itemize}
    \item While designed for global coherence, does BBScore capture both global and local coherence effectively in a synthetic setting? (Section \ref{exp:art})
    \item Can BBScore detects texts where the departure from desired coherence are not manipulated and recognize their differences ? (Section \ref{exp:down})
\end{itemize}

\subsection{Dataset}

In this paper, our focus is on the WikiSection dataset (city) as introduced in \citep{arnold-etal-2019-sector}. This dataset comprises Wikipedia articles related to global cities, sourced from Wikipedia dumps. Each segment of text within these articles is meticulously labeled with its corresponding section class and word count.

For our research, we make use of the pre-processed WikiSection data provided by \citet{wang2023language}. This curated dataset encompasses 2,165 articles in the training set and an additional 658 articles in the test set, all classified into 4 distinct section classes. Unlike the approach taken by \citet{wang2023language}, we deliberately exclude any section-related information during the training of the encoder. This decision is aimed at preventing the inadvertent disclosure of supplementary information and enhancing the model's overall applicability.

We then alter the orignal text to create material for each experiment. For the purpose of artificial tasks, we permute the original corpus to create various shuffled versions. As for downstream tasks, we employ large generative language models to complete each section. This is achieved by providing the model with the respective section title and the first few sentences as a prompt.

Additional details regarding the creation of the data are elaborated within the context of each specific experimental setup.

\subsection{Methods for Comparison}
We compare our proposed BBScore against the following coherence metrics:

\paragraph{Entity Grid} \citet{barzilay-lapata-2005-modeling} is the most recognized entity-based approach. For each input document, it creates a two-way contingency table to track the entity appearance in each sentence. The issue with this count-based approach is the grids can easily become sparse as the length of the document increases or when the text is informal \citep{lai-tetreault-2018-discourse}. The algorithm struggles to identify meaning entity transitions across sentences. This issue persists even in the subsequent neural version of Entity Grid. We include these methods in the comparison as the benchmark to validate the efficacy of BBScore. We use Stanford's CoreNLP to annotate the documents and the implementation in coheoka library\footnote[1]{\url{https://github.com/kigawas/coheoka}} to obtain the Entity Grid score.
\paragraph{Unified Coherence} \citet{moon-etal-2019-unified} is a neural-based entity-grid method that incorporates sentence grammar, inter-sentence coherence relations, and global coherence patterns into a common neural framework, and it reaches state-of-the-art results in many artificial tasks. However, this model architecture is trained end-to-end with a pairwise ranking loss, and in our opinion, these pairwise constraints can help the model reach a good performance in some tasks, but limits its application and generalization, since in the real world not all the incoherent article is an exact shuffled version of a coherent one. 
\paragraph{BBScore} Figure \ref{Figure:score_computation} shows a framework for the computation of BBScore. For example, in the WikiSection task, we train the encoder with the original training dataset and approximate the diffusion coefficient with the latents of the training dataset. The score is supposed to represent the deviation from the latent Brownian trajectory from training data.
\paragraph{BBScore + Classifier}\label{bbclf} We use global BBScore and local moving window BBScore with window sizes (defined in Appendix \ref{sec:app2_bb_window}) being 1, 2, 4, and 8 as features showing how coherent the text is. A three-layer perceptron then takes those features as input and outputs predictions for designed tasks.

\subsection{Artificial Tasks}\label{exp:art}

\paragraph{Global Discrimination} The evaluation of coherence can be effectively assessed through the Shuffle test \citep{barzilay-lapata-2008-modeling, moon-etal-2019-unified}. This evaluation method involves randomly permuting sentences within a discourse to create an incoherent document and compared to the original one. By employing this task, the overall coherence of the text can be effectively evaluated. To ensure the robustness and sensitivity of our proposed coherence scoring mechanism, we also incorporated the k-Blocked Shuffle test, which introduces varying levels of incoherence into the documents. In this test, we first trained the Brownian encoder using the WikiSection training set and then mapped both the original and shuffled documents into the latent space, generating their respective BBScores. 

To create the dataset, we generated a single random shuffle for each document and constructed a mixture of raw and shuffled documents, forming the basis of a binary classification task. We used BBScores to discern between coherent and incoherent text without additional supervision. To further align with established methodologies \citep{joty-etal-2018-coherence, moon-etal-2019-unified, jeon-strube-2022-entity}, we extended our experimentation to encompass a larger number of shuffles. Specifically, we generated 20 random shuffles for each document, both within the training and test sets. Furthermore, to maintain the integrity of the evaluation process, any documents that matched their original copy were omitted from the analysis. The statistics of the resultant dataset for 20 shuffle global discrimination tasks are shown in Table \ref{datastat}.

\begin{table}[t!]
  \centering
  \small
  \begin{tabular}{@{}ccc|cc}
    \toprule
    & \multicolumn{2}{c|}{Global} & \multicolumn{2}{c}{Local} \\
    \cmidrule(lr){2-3} \cmidrule(lr){4-5}
    & Data & \#Pairs & Data & \#Pairs \\
    \midrule
    \multirow{4}{*}{Train} & $\mathcal{D}_{b=1}$ &  36,574 & $\mathcal{D}_{w=1}$  &20,762 \\ 
    & $\mathcal{D}_{b=2}$ & 22,738  & $\mathcal{D}_{w=2}$ &26,165  \\ 
    & $\mathcal{D}_{b=5}$ & 17,110 & $\mathcal{D}_{w=3}$ &27,069 \\ 
    & $\mathcal{D}_{b=10}$ & 11,356 & $\mathcal{D}_{w=1,2,3}$ &73,996 \\
    \midrule
    \multirow{4}{*}{Test} & $\mathcal{D}_{b=1}$ &8,600 & $\mathcal{D}_{w=1}$  &6,255 \\ 
    & $\mathcal{D}_{b=2}$ &4,280 & $\mathcal{D}_{w=2}$  &7,924 \\ 
    & $\mathcal{D}_{b=5}$ &  3,837 & $\mathcal{D}_{w=3}$ &8,176 \\ 
    & $\mathcal{D}_{b=10}$ & 2,193 & $\mathcal{D}_{w=1,2,3}$  &22,355 \\
    \bottomrule
  \end{tabular}
  \caption{Statistics for the shuffle task dataset}
  \label{datastat}
\end{table}

\paragraph{Local Discrimination}\label{exp:local} Detecting incoherent documents in the global discrimination task is relatively straightforward; however, assessing local coherence at the document level presents a more challenging task. To evaluate the effectiveness of our coherence scoring approach for local coherence, we constructed datasets using local permutation windows.

Following the methodology outlined in \citet{moon-etal-2019-unified}, we randomly sampled a set of windows, denoted as $w$, from the entire text. Each window consisted of a fixed size of three sentences, ensuring there was no overlap between any of the windows. Subsequently, we randomly shuffled the order of sentences within each window. We repeated the shuffle procedure for 20 times, excluding any duplicate sample.

\begin{figure}[t!]
  \centering
  \includegraphics[width=\linewidth,trim={2.9cm 0 0 3.5cm},clip]{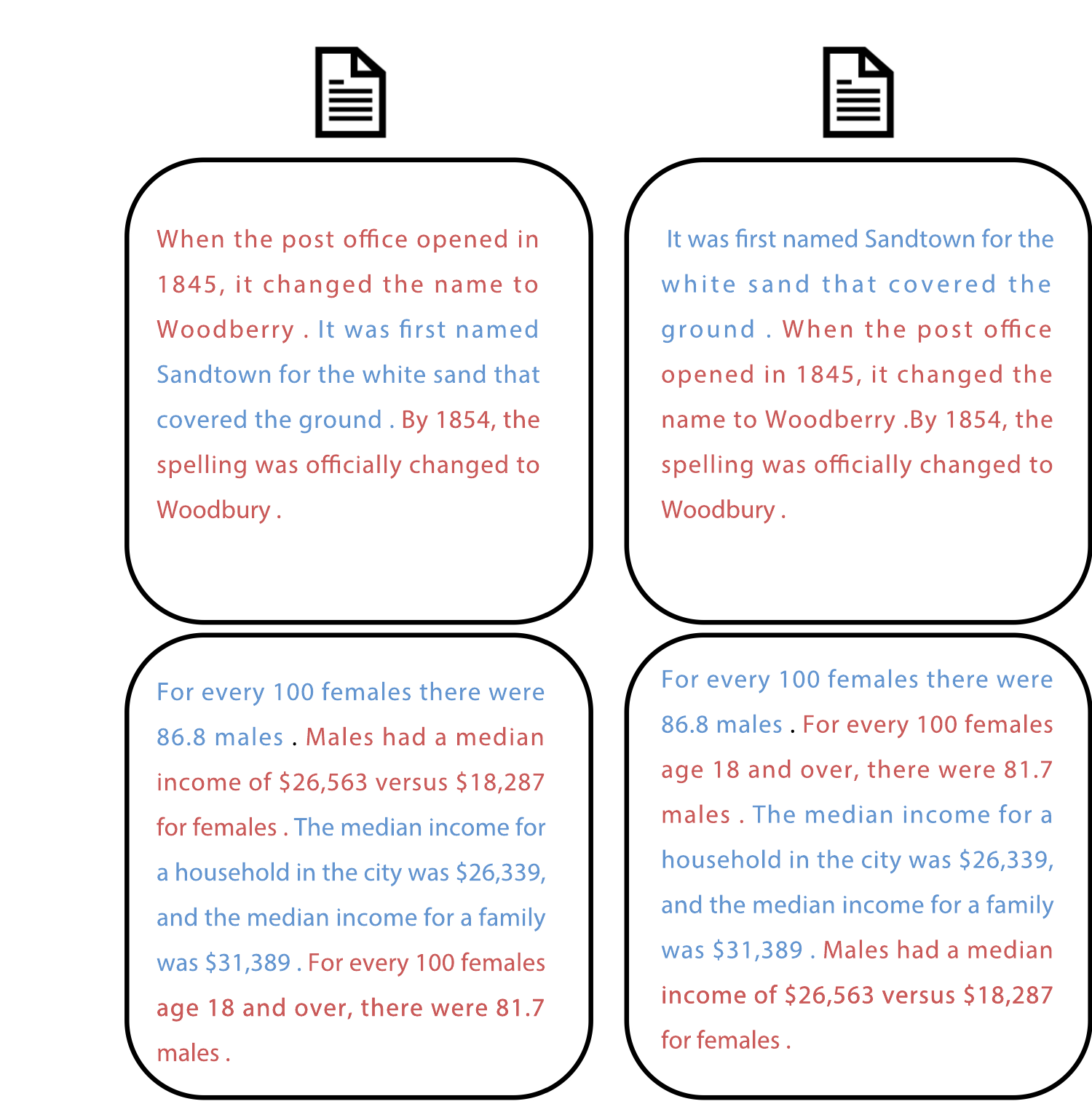}
  \caption{Example window pair for local discrimination task. \textbf{Left} is from the locally shuffled article and the \textbf{right} is from the original article.}
  \label{Figure:local_exp}
\end{figure}

We generated three separate datasets for the local discrimination task, namely $\mathcal{D}_{w=1}$, $\mathcal{D}_{w=2}$, and $\mathcal{D}_{w=3}$. Each dataset had a different window size, which determined the degree of evaluation for local coherence. Figure \ref{Figure:local_exp} demonstrates examples of incoherence caused by local shuffles.

\subsection{Downstream Task}\label{exp:down}
As noted in a previous study \citep{mohiuddin-etal-2021-rethinking}, there is a limited correlation between model performance on artificial tasks and their performance on downstream tasks. Since BBScore is not restricted by a well-defined structure, such as pairwise input, we further explore its potential application areas. We assess our method on two novel downstream tasks: a task aimed at distinguishing LLM-generated articles from human-written ones and another task to detect which LLM the given input texts are from.

\paragraph{AI Differentiation} For this task, we use the WikiSection dataset as human-written texts. To obtain LLM-generated texts with similar structures as the human-written ones, we use the first few sentences in each section of the WikiSection article as prompts, ask LLM to complete each section, and then combine those sections into one generated article. We do not compare to \citet{moon-etal-2019-unified} in this task because it requires the text length to be aligned between the pair. We still include the Entity Grid as the baseline.

\paragraph{LLM Detection} In order to exam the potential generalizability of BBScore, we form a domain detection task where texts generated by different large language models can be viewed as different domains based on our assumption. Our primary inquiry revolves around the sensitivity of $\hat\sigma^2_m$ in distinguishing the employed large language model during the text generation process when provided with input texts.

\section{Results}
\label{sec:results}

\begin{table*}[ht]
\centering
\small
\resizebox{1.0\linewidth}{!}{%
\renewcommand{\arraystretch}{1.2}
\begin{tabularx}{\textwidth}{>{\small\arraybackslash}p{5.5cm}|  *{8}{>{\centering\arraybackslash}X}}
\toprule
\multirow{2}{*}{\textbf{Methods}} & \multicolumn{4}{c}{\textbf{Train}} & \multicolumn{4}{c}{\textbf{Test}} \\
\cmidrule(lr){2-5} \cmidrule(lr){6-9}
& $\mathcal{D}_{b=1}$ & $\mathcal{D}_{b=2}$ & $\mathcal{D}_{b=5}$ & $\mathcal{D}_{b=10}$ & $\mathcal{D}_{b=1}$ & $\mathcal{D}_{b=2}$ & $\mathcal{D}_{b=5}$ & $\mathcal{D}_{b=10}$ \\
\midrule
\textsc{EntityGrid} \cite{barzilay-lapata-2008-modeling} &79.17  &86.20  &74.51  &60.70  &85.73  &82.79 &75.81  &64.65  \\
\textsc{UnifiedCoherence} \cite{moon-etal-2019-unified} & \textbf{99.75} & 98.60 & 97.10 & 96.21 & \textbf{99.73} & 97.86 & 96.90 & 96.09 \\
\textsc{BBScore} & 76.29 & 75.12 & 73.04 & 73.12 & 83.39 & 80.71 & 79.36 & 78.66 \\
\textsc{BBScore + CLF} & 99.12 & \textbf{98.95} & \textbf{98.68} & \textbf{98.54} & 98.92 & \textbf{98.46}& \textbf{98.25} & \textbf{98.50} \\

\bottomrule
\end{tabularx}%
}
\caption{Global Discrimination Task Results on WikiSection}
\label{wiki-global}
\end{table*}

\begin{table*}[htbp]
  \centering
  \small
  \resizebox{1.0\linewidth}{!}{%
  \renewcommand{\arraystretch}{1.3}
  \begin{tabularx}{\textwidth}{>{\small\arraybackslash}p{5.5cm}| *{8}{>{\footnotesize\centering\arraybackslash}X}}
    \toprule
    \multirow{2}{*}{\textbf{Methods}} & \multicolumn{4}{c}{\textbf{Train}} & \multicolumn{4}{c}{\textbf{Test}} \\
    \cmidrule(lr){2-5} \cmidrule(lr){6-9}
    & $\mathcal{D}_{w=1,2,3}$ & $\mathcal{D}_{w=1}$ & $\mathcal{D}_{w=2}$ & $\mathcal{D}_{w=3}$ & $\mathcal{D}_{w=1,2,3}$ & $\mathcal{D}_{w=1}$ & $\mathcal{D}_{w=2}$ & $\mathcal{D}_{w=3}$ \\
    \midrule
    \textsc{EntityGrid} \cite{barzilay-lapata-2008-modeling} &61.25  &55.37  &62.94  &65.44  &60.18  &53.04  &60.83 &66.67 \\
    \textsc{UnifiedCoherence} \cite{moon-etal-2019-unified}&90.84 &84.02 &83.64 &89.59 &87.00 &77.47 &82.98 &87.87  \\
    \textsc{BBScore}&57.85 & 47.17 & 55.84 & 63.10 &57.66 &50.29 & 60.15 & 64.06  \\
    \textsc{BBScore + CLF}& 69.68 & 61.69 & 69.00 & 75.04 & 67.12 & 55.20 & 67.86 &  75.66   \\

    \bottomrule
  \end{tabularx}%
  }
\caption{Local Discrimination Task Results on WikiSection.  $\mathcal{D}_{w=1,2,3}$ represents the joint set of all three other datasets.}
\label{wiki-local}
\end{table*}

\subsection{Global Discrimination Task}
\begin{figure}[t!]
  \centering
  \hspace*{0mm}
  \begin{subfigure}[b]{0.48\linewidth}
    \centering
    \includegraphics[width=\linewidth,trim={0.2cm 0.8cm 1cm 1cm},clip]{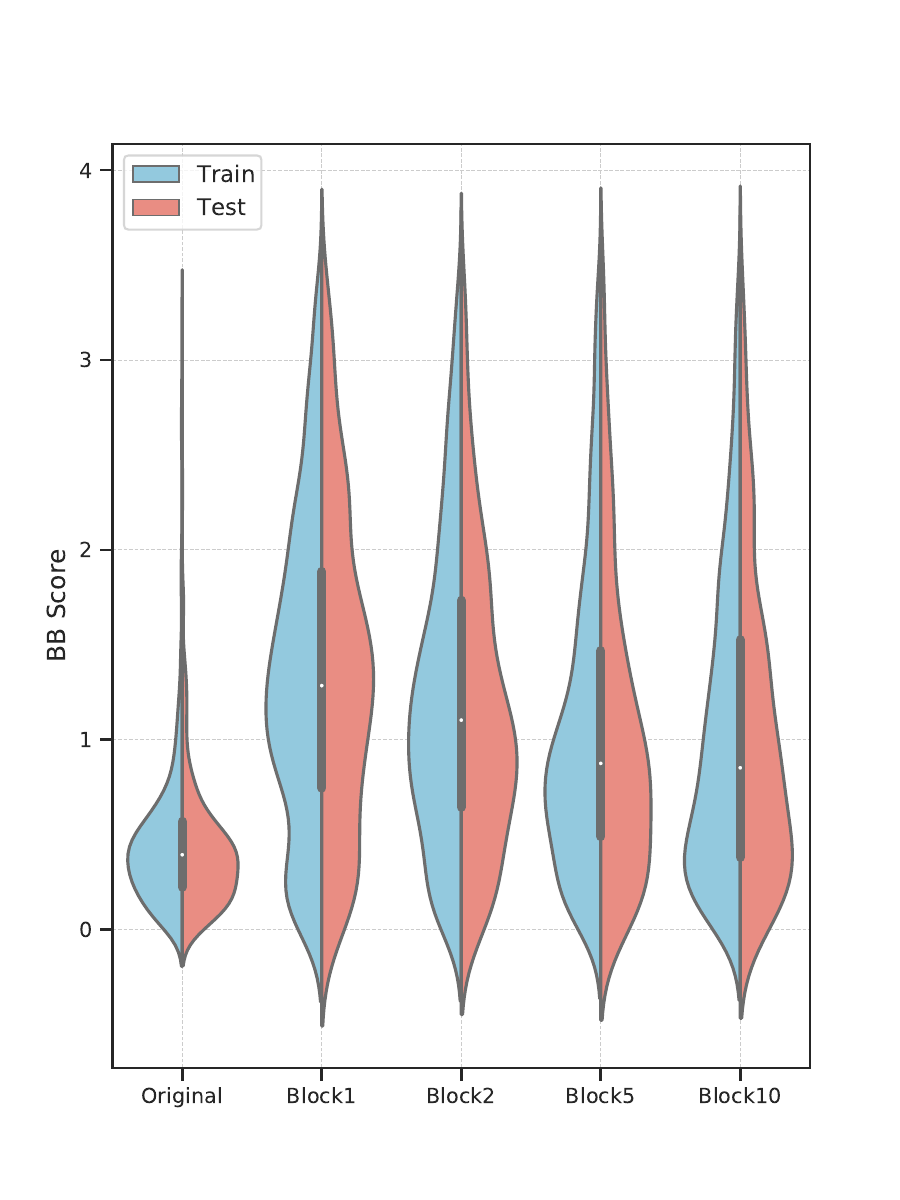}
    \label{violin_bb}
  \end{subfigure}
  \hspace{0.3mm}
  \begin{subfigure}[b]{0.48\linewidth}
    \centering
    \includegraphics[width=\linewidth,trim={0cm 0.8cm 1cm 1cm},clip]{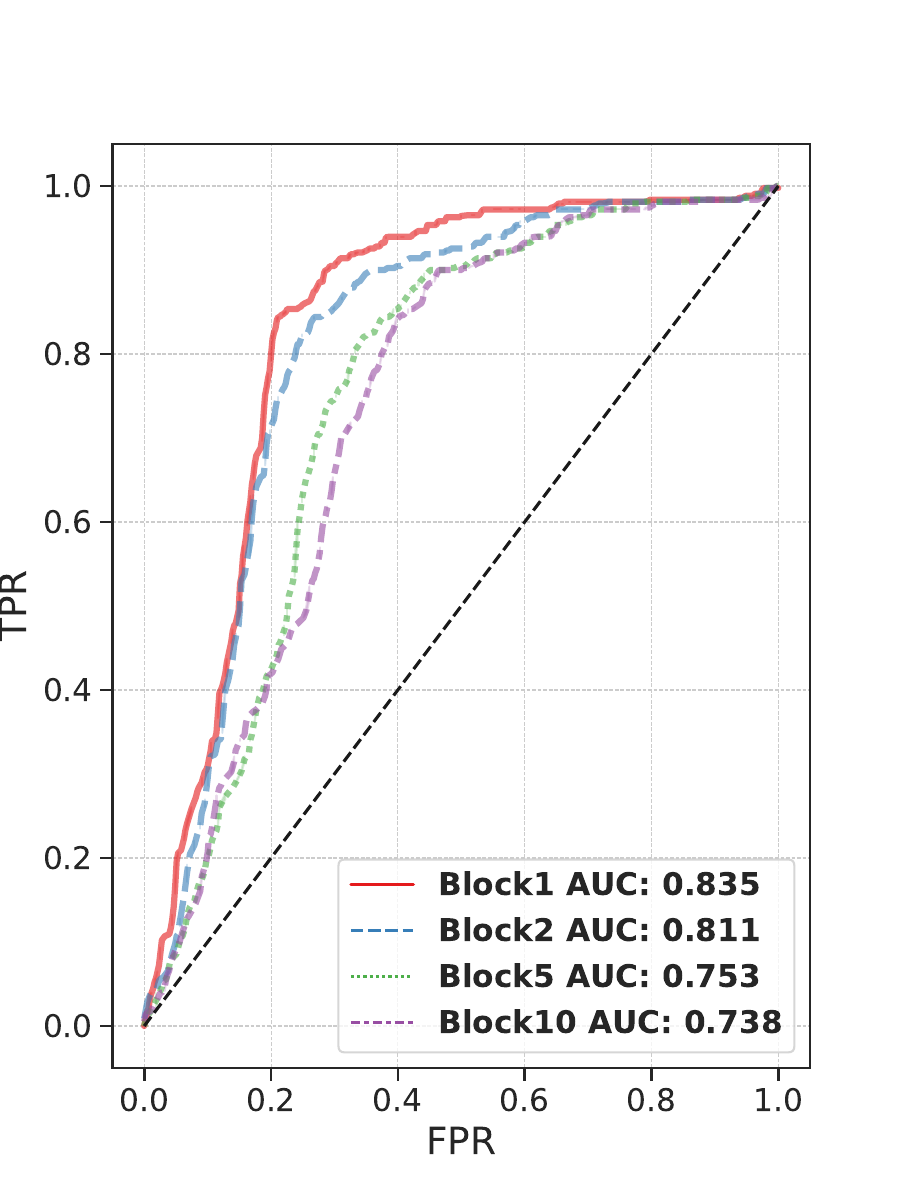}
    \label{auc_test}
  \end{subfigure}
  \caption{\textbf{Left}: BBScore distribution with different size of shuffle blocks. \textbf{Right}: Test set BBScore-based AUC with different size of shuffle blocks}
   \label{fig2}
\end{figure}

We calculate the BBScore for the original training and test set as well as the ones with various shuffle windows, and from their distribution we can conclude coherent text ought to have a lower BBScore, which implies that they are closer to the hypothetical Brownian bridge process. We also rank the scores from all the articles and compute the AUC score for coherent and incoherent separation.
The general performance of our model on shuffled tasks is shown in Figure \ref{fig2}. The AUC for Block 1 shuffle reaches 0.835 with the raw BBScore and remains at 0.738 for Block 10 shuffle where the incoherence is harder to detect.

Next, we calculate the pairwise accuracy score for each document with scores derived from different methods, and the result is shown in Table \ref{wiki-global}. 
As the block size increases in the block shuffle task, the task becomes more challenging. This is because the shuffled article exhibits greater local coherence, but overall, it still lacks sufficient global coherence. The performance of the basic Entity Grid score on the test set shows significant variation with changes in block sizes. Consequently, to successfully accomplish this task, the metric should consider both local and global coherence while maintaining a balance between them.

From the result of the basic BBScore, we can find that with an encoder trained on the original Train dataset and diffusion coefficient approximated with latents of the Train dataset, its performance is still very stable even if the task becomes harder. After we add a simple linear classifier after the BBScore, its performance exceeds \cite{moon-etal-2019-unified} and becomes better when the block size increases. Therefore, we conclude the BBScore has a better evaluation of global coherence while still capturing local coherence. To further explore and verify the ability of BBScore on local coherence evaluation, we implement the local coherence discrimination task in the following section.

\subsection{Local Discrimination Task}
We compute pairwise scores for each document and its locally shuffled counterpart. The accuracy reported is based on whether the model can assign a lower BBScore to the original document compared to the locally shuffled one in each pair. The results can be found in Table \ref{wiki-local}. As expected, the Unified Coherence Model \citep{moon-etal-2019-unified} demonstrates strong performance across different locally shuffled datasets, given that their model incorporates sentence alignment and deliberately learns to differentiate between document pairs. Their model includes a dedicated module for capturing inter-sentence relations and modeling local coherence. In contrast, our approach focuses more on global coherence rather than supervising local windows. Nonetheless, our supervised model, BBScore+CLF, demonstrates the fundamental ability to evaluate local coherence.

\subsection{AI Generated Text Differentiation}
In this section, we mainly focus on two different training strategies: (1) Pairwise discrimination; and (2) General AI discrimination. 

For the pairwise discrimination task, we design a data pair, e.g. (original doc, language model generated doc) as an input to test whether we can distinguish which one was generated by humans. 
An example is shown in Appendix \ref{pairai}, and in this pair, the document from WikiSection has a smaller BBScore than the AI-generated document, so we can successfully detect the AI-generated one. Later, we also train a simple classifier with the BBScore on this pairwise classification task, and the complete result is listed in Table \ref{tab:ai_diff}, where we use six different large language models to generate the document.

\begin{figure}[ht]
\centering
\hspace*{0mm}
\begin{subfigure}[b]{0.48\linewidth}
\centering
\includegraphics[width=\linewidth,trim={0cm 0.8cm 1cm 1cm},clip]{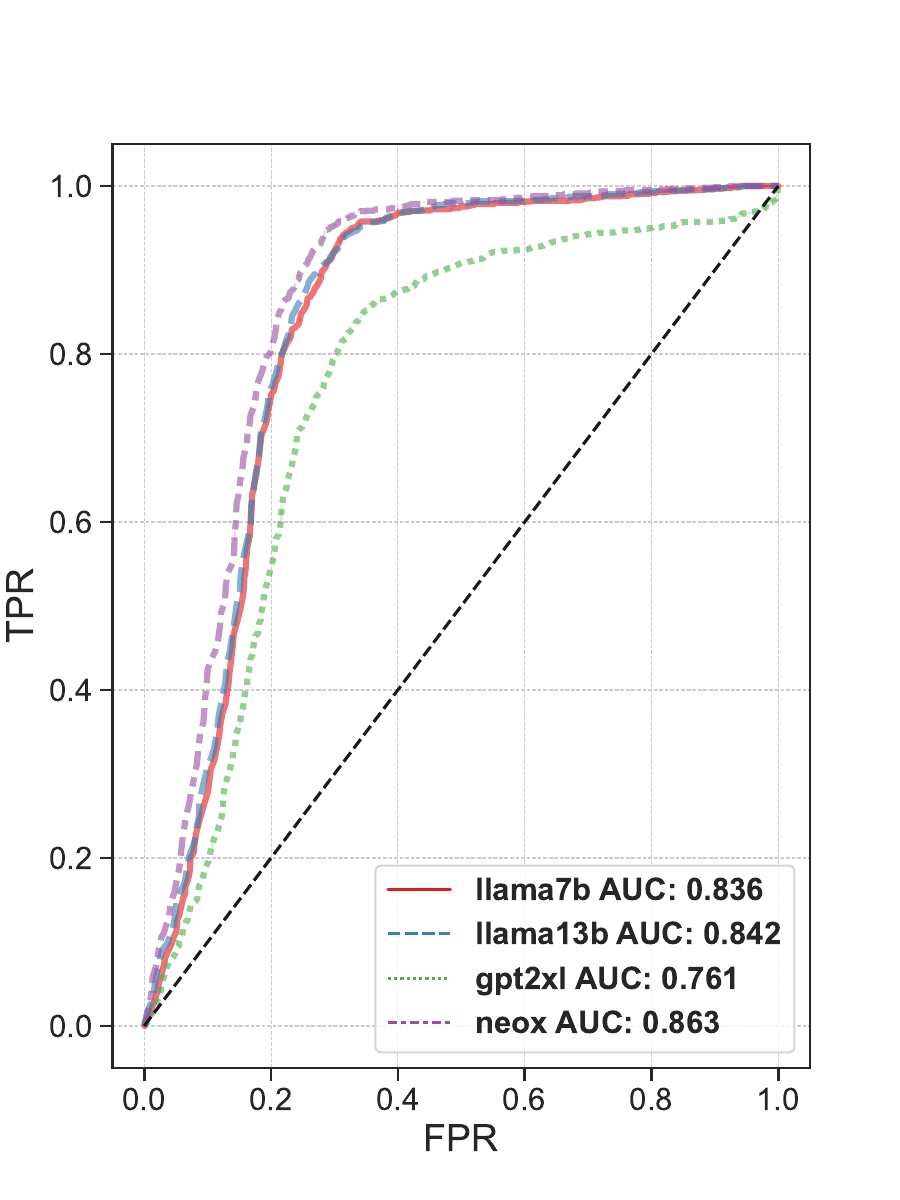}
\label{ai_auc_train}
\end{subfigure}
\hspace{0.3mm}
\begin{subfigure}[b]{0.48\linewidth}
\centering
\includegraphics[width=\linewidth,trim={0cm 0.8cm 1cm 1cm},clip]{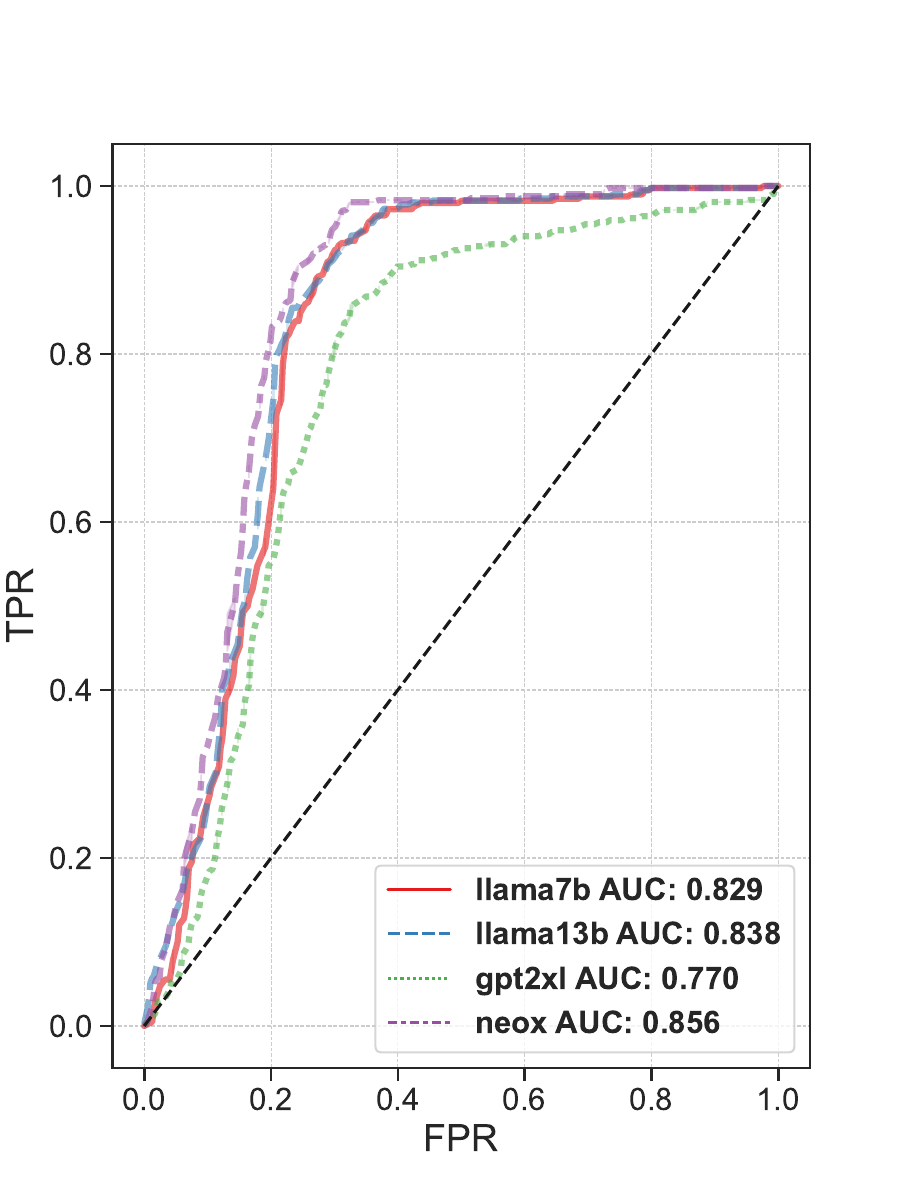}
\label{ai_auc_test}
\end{subfigure}
\vspace{-1cm}
\caption{\textbf{Left} and \textbf{Right} correspond to \textbf{Train} and \textbf{Test} BBScore-based AUC with different LLMs, respectively.}
\label{ai_auc}
\end{figure}

General AI discrimination task is a harder problem. With any document as an input, we train BBScore+CLF to predict whether humans or the AI write this document, the classification accuracy are shown in Table \ref{tab:ai-general}. We also compute the AUC score as shown in Figure \ref{ai_auc}. The AUC corresponds to different large language models that do not vary a lot which implies our model can finish the general AI discrimination task in a stable way.

\begin{table*}[t!]
\centering
\small
\resizebox{1.0\linewidth}{!}{%
\renewcommand{\arraystretch}{1.3}
\begin{tabularx}{\textwidth}{>{\small\arraybackslash}p{2.4cm}| *{12}{>{\centering\arraybackslash}X}}
\toprule
\multirow{2}{*}{\textbf{Methods}} & \multicolumn{6}{c}{\textbf{Train}} & \multicolumn{6}{c}{\textbf{Test}} \\
\cmidrule(lr){2-7} \cmidrule(lr){8-13}
& $\clubsuit$ & $\spadesuit$ & $\vardiamondsuit$ &  $\varheartsuit$ & $\diamondsuit$ & $\heartsuit$
& $\clubsuit$ & $\spadesuit$ & $\vardiamondsuit$ &  $\varheartsuit$ & $\diamondsuit$ & $\heartsuit$\\
\midrule
\textsc{EntityGrid} &54.43  &36.99  &18.64 &25.23 &43.50 &25.86 &59.57 &36.17 &25.88 &32.19 &39.40 &22.85  \\
\textsc{BBScore} &76.96  &85.07 &82.21 &82.42 &73.07 &75.29 &77.75  &83.74 &81.16 &81.82 &75.63 &75.13\\
\textsc{BBScore + CLF} & \textbf{83.21} & \textbf{85.51} & \textbf{88.04}& \textbf{88.03} &\textbf{92.19} &\textbf{91.89} & \textbf{83.01} &\textbf{86.17} &\textbf{86.18} & \textbf{86.73} &\textbf{92.02} &\textbf{91.89} \\
\bottomrule
\end{tabularx}%
}
\caption{AI Generated Text Differentiation Results on WikiSection. $\clubsuit$: GPT2XL $\spadesuit$: GPT-NeoX $\vardiamondsuit$: LLaMA-7B $\varheartsuit$: LLaMA-13B $\diamondsuit$: LLaMA2-7B $\heartsuit$: LLaMA2-13B}
\label{tab:ai_diff}
\end{table*}

\begin{table}[t!]
  \centering
  \renewcommand{\arraystretch}{1.2} 
  \footnotesize
  \begin{threeparttable}[b]\    
  \label{gai}
  \begin{tabularx}{\linewidth}{>{\small\arraybackslash}p{1cm}|*{6}{>{\centering\arraybackslash}X}}
    \toprule
    \multirow{2}{*}{\textbf{Dataset}} & \multicolumn{6}{c}{LLM} \\
    & $\clubsuit$ & $\spadesuit$ & $\vardiamondsuit$ &  $\varheartsuit$ & $\diamondsuit$ & $\heartsuit$ \\
    \midrule
    \textbf{Train} & 77.33 & 83.42 & 84.55 & 82.80 & 85.12 & 86.80\\
    \textbf{Test} & 76.44 & 82.65 & 82.53 & 81.20 & 84.53 & 85.35\\
    \bottomrule
  \end{tabularx}
  \caption{General AI Discrimination Task Results of BBScore + CLF on WikiSection. $\clubsuit$: GPT2XL $\spadesuit$: GPT-NeoX $\vardiamondsuit$: LLaMA-7B $\varheartsuit$: LLaMA-13B $\diamondsuit$: LLaMA2-7B $\heartsuit$: LLaMA2-13B}
\label{tab:ai-general}
\end{threeparttable}
\end{table}

\subsection{LLM Detection}
\begin{figure}
    \centering
    \includegraphics[width=0.9\linewidth, trim={3cm 1cm 4cm 2cm},clip]{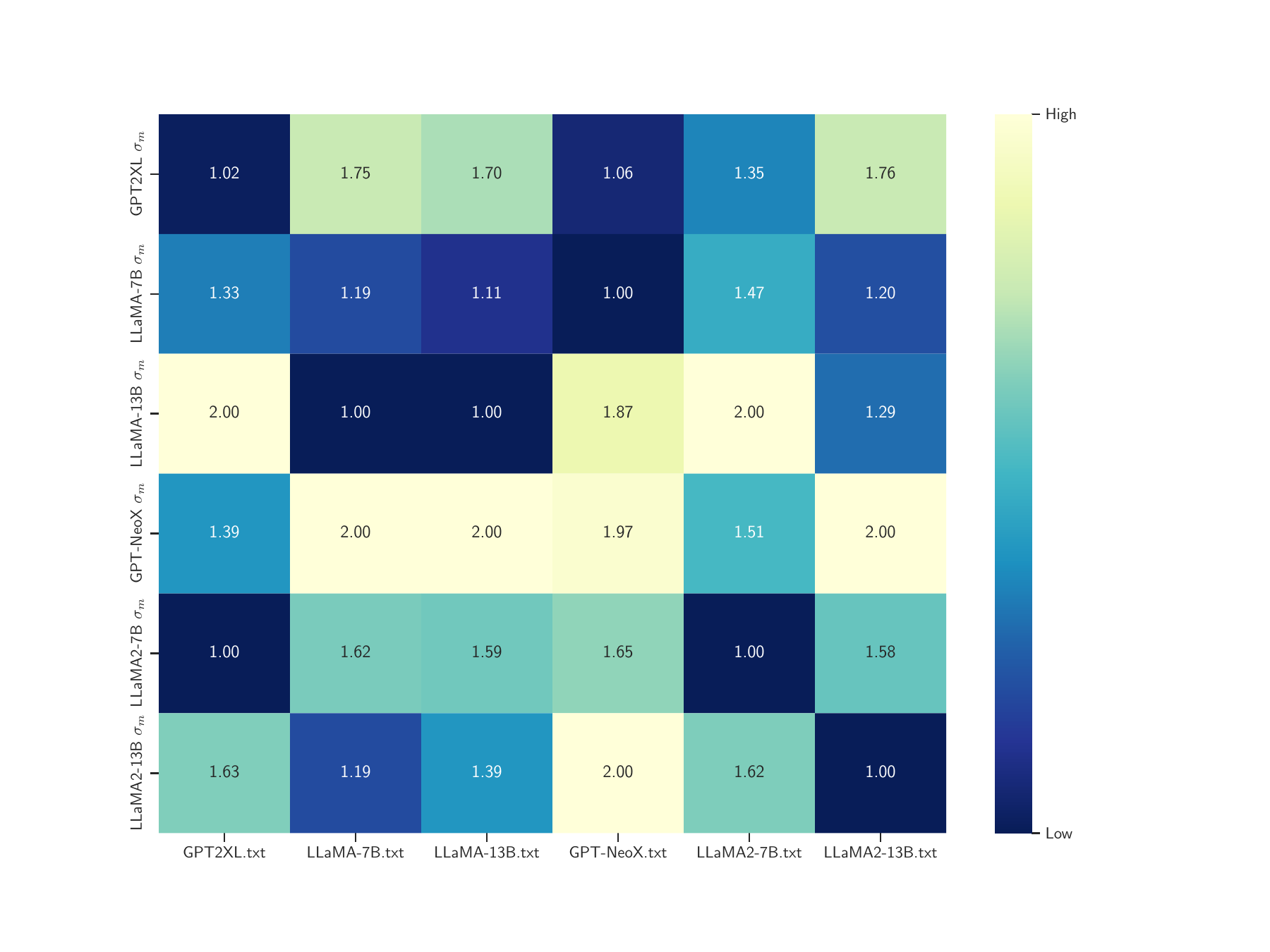}
    \caption{Normalized Wasserstein Distances. Each column represents the input texts are generated by a specific model.}
    \label{emover}
\end{figure}

For this task, we initiate by training a Brownian encoder utilizing a composite corpus consisting of articles generated by diverse LLMs and the original articles. Subsequently, we derive estimates for $\sigma^2_{m_k}$ from the training set corresponding to LLM $k$. Then we calculate $\sigma^2_{m}(\ss)$ conditional on the input text $\ss$ obtained from the test set. Moving forward, we evaluate the distributional similarity between $\sigma^2_{m}(\ss)$ and $\sigma^2_{m_k}$, and normalize the outcomes for each $\ss$ to ensure the values are on the scale from 1 to 2. The outcomes, as calculated by the Wasserstein distance, are illustrated in Figure \ref{emover}.
In the results, we observe that 5 out of the 6 test texts are in close proximity (among the top 2) to their respective source LLMs, except for the text generated by GPT-NeoX. This illustrates that the comparison of the distribution of $\sigma_m^2$ effectively distinguishes a narrower range of LLMs to be considered. Given that the input texts pertain to the same subject and variations introduced by different LLMs can be subtle, we find this technique valuable in identifying subdomains in the condition of training with an extensive collection of text and can potentially be generalized into other domains. To examine the generalizability in terms of out of domain performance, we also test the WikiSection encoder on the GCDC dataset \citep{lai-tetreault-2018-discourse} (See Appendix \ref{app:gcdc}). 

\subsection{Trajectory in Latent Space}
\begin{figure}[htbp]
    \centering
    \includegraphics[width=\linewidth, trim={3cm 0cm 3cm 0cm},clip]{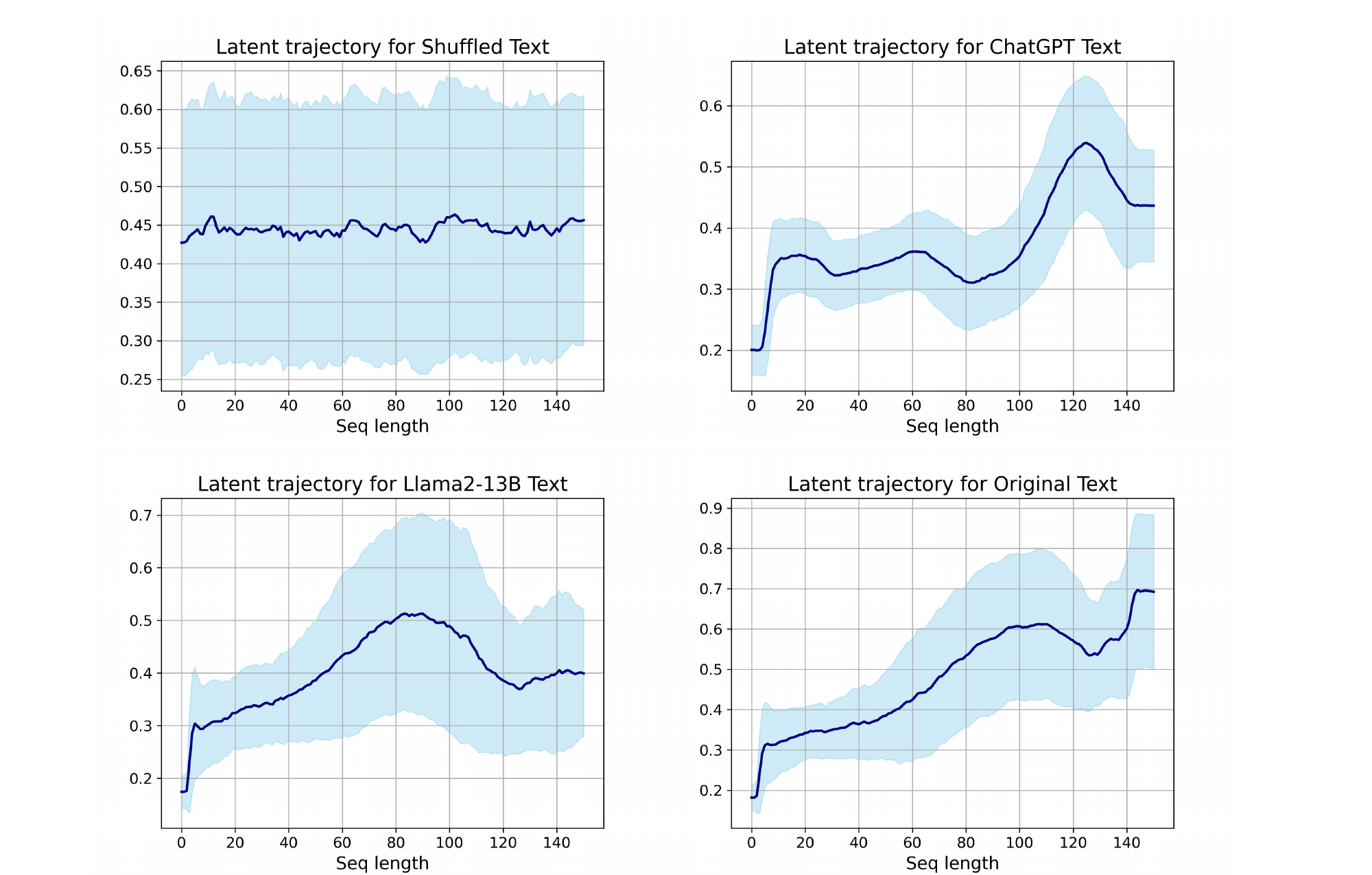}
    \caption{Latent trajectories for text from different generation processes.}
    \label{text_latents}
\end{figure}

In this section, we visualize the trajectories in the latent space of texts from different generation processes mapped by the trained Brownian encoder. Our goal is to examine whether the assumption about latent coherence, as depicted in Figure \ref{Figure:latents_flow}, holds in general. We test text from unseen WikiSection data, text generated by two large language models - ChatGPT and LLaMA2-13B, and text created by sentence-level permutation. The results are shown in Figure \ref{text_latents}. The latents are interpolated to uniform length of documents and then averaged over all documents at each sentence index. From the visualization, we observe that randomly shuffled text follows a flat trajectory where sentences are not in logical order, and similar sentences could occur at any location within the document, thus disrupting the flow of the text. Conversely, text written by humans or generated by LLMs tends to follow a trajectory more akin to a Brownian bridge in the latent space. The movement of the latents typically starts from lower values with a jump and then progresses to higher values, accompanied by fluctuations in the middle. Concurrently, we also notice that the variance of the latents initially increases and then diminishes towards the end, particularly in human-written text. This observation aligns with the properties of a Brownian bridge process. These findings validate our assumption that coherent text resembles a Brownian bridge in the latent space, as per our modeling process.

\section{Discussion and Limitations}
In this paper, we propose the BBScore as a simple yet effective metric for measuring text coherence. Unlike previous entity-based approaches that rely on linguistic insights, the BBScore naturally learns text coherence through a contrastive objective during training. According to \cite{grosz-sidner-1986-attention}, three factors contribute to discourse coherence: the organization of discourse segments, the intention or purpose of the discourse, and attention or focused items. While well-organized discourse segments are common in formal writing, the abundance of informal text poses challenges in conveying the main topic or intention through entity transitions. In such cases, explicit guidance for learning text coherence becomes more difficult, necessitating a self-supervised approach. 

In Section \ref{bbclf}, we demonstrate that BBScore could be calculated at the sentence level, and by manipulating the sentence window sizes, we can generate a combination of features that capture coherence at different discourse units. Despite its simple network structure, BBScore achieves state-of-the-art results on global discrimination tasks in standard coherence assessment experiments, highlighting its potential for evaluating global text coherence. Additionally, even without a specific structure for capturing local coherence, our model still performs well in evaluating local text coherence. We believe that the diffusion coefficient encodes important domain-specific features that aid in text coherence evaluation. More specifically, we show it can be leveraged for Human-AI discrimination and used to compare generation qualities of different models with respect to coherence. From the LLM detection task results, we find when training with an inclusive corpus, $\hat\sigma_m$ can potentially be utilized to identify sub-domains. 
Our results also demonstrate that while Entity Grid is a simple and effective measure in artificial settings, it falters in downstream tasks.

While our approach has yielded positive results, we also identify several limitations. In this paper, our focus has been on a single dataset to showcase the usability of BBScore. Although we have demonstrated its ability to discern different LLM-generated content by training with a cross-domain corpus, the obtained results are not entirely satisfactory and still require further robustness validation.
In light of this, we acknowledge that the assumption we have made, linking $\sigma_m^2$ to specific domain/style, is rather restrictive in its general applicability. Future endeavors will involve expanding the parameter space by introducing higher-dimensional variance estimates. Furthermore, given BBScore's demonstrated capacity to differentiate between LLM-generated text and human-authored text, we aspire to establish BBScore as a more general and well-defined metric for comparing various LLMs in our forthcoming research. Lastly, BBScore's correlations with formal Human evaluation should be examined. 

\section{Conclusion}
Overall, BBScore presents a novel perspective on text coherence and has demonstrated its efficacy on artificial tasks involving deliberately induced incoherence. Additionally, we illustrate the practical utility of BBScore in a natural context, where unintentional deviations from desired coherent text occur. Serving as an intermediate computed score, BBScore holds the potential to become a valuable feature in numerous real-world applications, including tasks related to Human-AI discrimination.
In contrast to the intricate network architectures employed in neural entity-grid models (e.g., Multi-layer LSTM), our approach utilizes a simple three-layer perceptron with BBScore as input for classification tasks, devoid of any crafted loss function. Remarkably, the experiments 
 show this approach attains comparable, and in some cases, even superior results. Further refinement of our existing method promises to be an intriguing avenue for future research.

\fontsize{9.5pt}{10.5pt}
\selectfont

\bibliography{aaai24}

\newpage
\begin{center}
\Large \textbf{Appendix}
\end{center}
\appendix
\section{Example of pairwise discrimination on AI discrimination tasks}
\label{pairai}
In the pairwise discrimination on AI discrimination tasks, for each data pair (original doc, AI-generated doc), we will compute the BBScore separately as we did in the coherence task, the local/global discrimination task. An example is shown in Figure S2, here the AI-generated document uses the first sentence of the original doc as a prompt and generates the rest document with LLaMA7b. In this example, the original document has a lower BBScore and the task on this data pair is marked as successful.
\begin{figure}[htbp]
\centering
\includegraphics[width=1.15\linewidth,trim={1.1cm 0cm 0cm 0cm},clip]{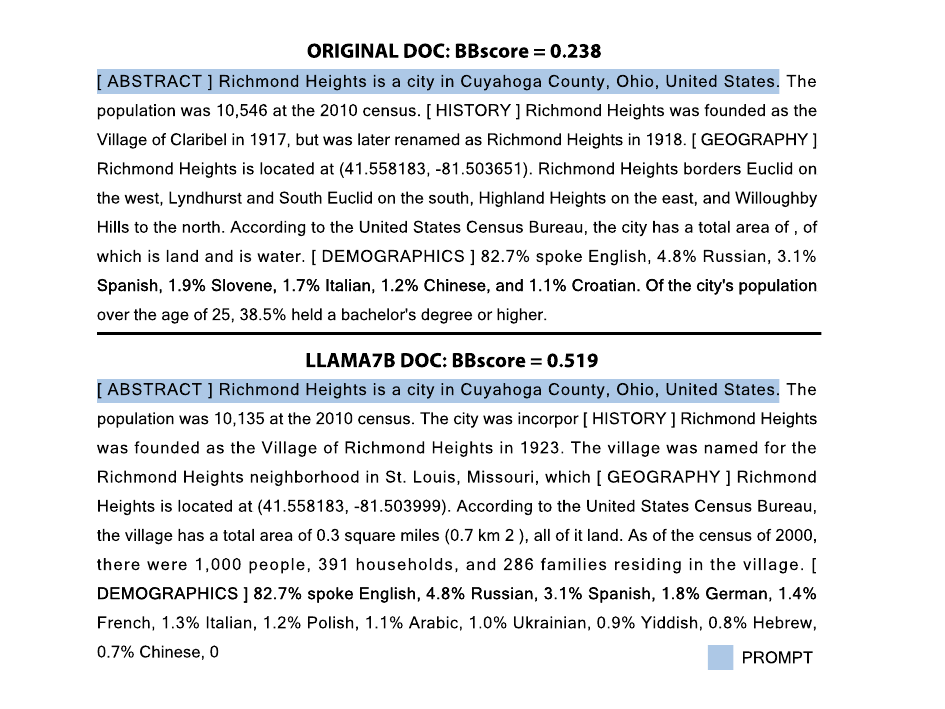}
\caption{An example of the pairwise AI discrimination}
\label{fig:ai_example}
\end{figure}

\section{Domain generalizability of Brownian encoders}
\label{app:gcdc}
We use the WikiSection encoder to encode GCDC texts and obtain the corresponding BBscores. A three-layer perceptron is then trained on the BBscores for a three-class classification task on the GCDC dataset. The results are shown in Table S1.

\begin{table}[htbp]
  \centering
  \renewcommand{\arraystretch}{1.3} 
  \footnotesize
  \begin{threeparttable}[b]\    
  \label{gcdc}
  \begin{tabularx}{0.9\linewidth}{>{\small\arraybackslash}p{3cm}|*{6}{>{\centering\arraybackslash}X}}
    \toprule
    \multirow{2}{*}{\textbf{Dataset}} & \multicolumn{4}{c}{\textbf{Domain}} \\
    & Enron & Clinton & Yahoo &  Yelp  \\
    \midrule
    \textbf{Train} & 47.67 & 43.11 & 49.54 & 51.45 \\
    \textbf{Test} & 47.50 & 41.50 & 42.64 & 49.25 \\

    \bottomrule
  \end{tabularx}
  \caption{Three-classs Classification Task Results on GCDC Dataset with the WikiSection Encoder.}
\end{threeparttable}
\end{table}

\section{Diffusion coefficient $\hat{\sigma}^2_m$ analysis}
\label{sec:app1_DCA}
As shown in Figure S1, it describes the AUC score for the blocksize=1 shuffle test under different diffusion coefficients. It shows our current approximation of the diffusion coefficient (marked by the red dashed line) can give us a better result but not the best (marked by the olive dashed line). Moreover, it also shows, the standard Brownian bridge $\sigma^2_m$=1 shows a poor result AUC score=1 which emphasizes the necessity of this diffusion coefficient approximation. 
\begin{figure}[htbp]
\centering
\includegraphics[width=0.7\linewidth, trim={1.1cm 3cm 0cm 2cm},clip]{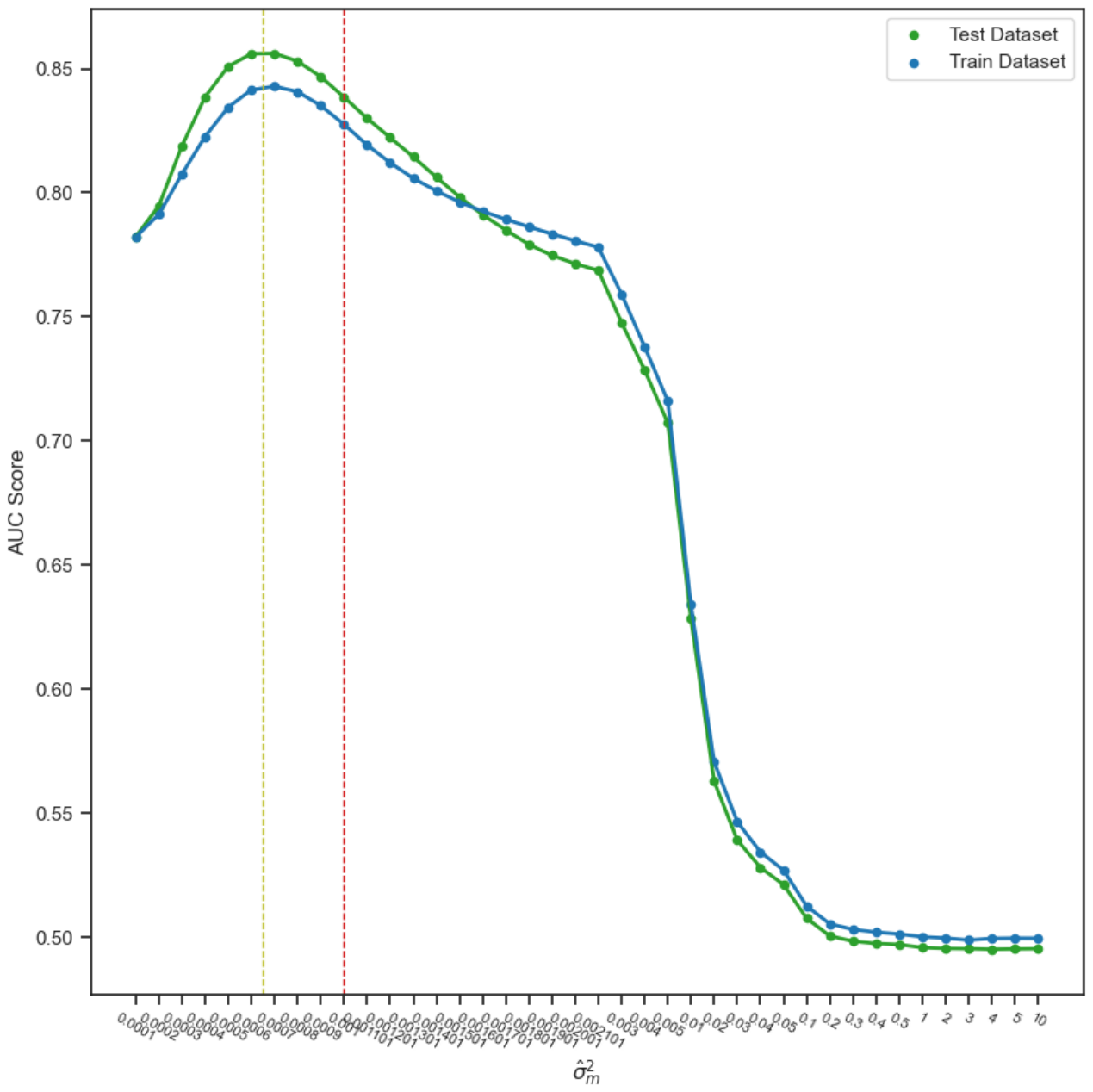}
\caption{Diffusion coefficient analysis: The AUC score for shuffle test with block 1 under different $\sigma^2_m$, and the red dash line corresponds to the $\hat{\sigma}^2_m$ approximated with the train dataset.}
\label{fig:diff_coeff_analysis}
\end{figure}

\section{BBScore defined with a shifting window}
\label{sec:app2_bb_window}
The basic BBScore is defined as:
\begin{equation}
\label{eq:evaluation_measure1}
\footnotesize
\begin{aligned}
\bb (\ss| \hat{\sigma}_m^2) = \frac{|\sum_{i=2}^{T(\ss)-1} \ln(\alpha_i(\ss) \hat{\sigma}_m^2)+ \frac{\beta_i(\ss)}{\hat{\sigma}_m^2}|}{T(\ss)-2}.
\end{aligned}
\end{equation}
We also test BBScore with a shifting window to capture the local coherence property: given a shifting window size $2w+1$, $w\in \mathbb{N}$, the shifting window BBScore $\bb_w (\ss| \hat{\sigma}_m^2)$ is defined as,
\begin{equation}
\label{eq:BBScore_shifting_windows}
\footnotesize
\begin{aligned}
\bb_w (\ss| \hat{\sigma}_m^2) = \frac{|\sum_{i=w+1}^{T(\ss)-w} \ln(\alpha_{i,w}(\ss) \hat{\sigma}_m^2)+ \frac{1}{\hat{\sigma}_m^2} \beta_{i,w}(\ss)|}{T(\ss)-2w}
\end{aligned}
\end{equation}
where for $i=w+1,\cdots T(\ss)-w$
\begin{equation*}
\footnotesize
\begin{aligned}
\alpha_{i,w}(\ss) = 2\pi \frac{w(w+1)}{2w+1}, \ \beta_{i,w}(\ss) = \frac{(2w+1)||s_i -\mu_i||^2}{ 2w(w+1)}
\end{aligned}
\end{equation*}
and
\begin{equation*}
\footnotesize
\begin{aligned}
\mu_i = s_{i-w} + \frac{w+1}{2w+1}(s_{i+w} - s_{i-w}).
\end{aligned}
\end{equation*}

\section{Training details}
\label{training_detail}
For all the experiments mentioned in the main paper, we used a hidden dimension size of 128 for the multi-layer perceptron appended to the GPT2 encoder and an output dimension size of 8 for the fully connected layer. The Brownian encoders were then trained using the contrastive objective function via the SGD optimizer, with learning rate of $1 \times 10^{-4}$, momentum of 0.9, and batch size of 32. The GPT2-based encoder was trained on 1 node with 2 A100 GPUs and 32 GB of memory.

\end{document}